\documentclass[conference]{IEEEtran}
\IEEEoverridecommandlockouts
\usepackage{graphicx}
\usepackage[utf8]{inputenc}
\usepackage{amssymb,amsmath,array}
\usepackage{multirow}
\usepackage{multicol}
\usepackage{booktabs}
\usepackage{url}

\usepackage{tikz}
\usetikzlibrary{decorations.pathmorphing}

\usepackage{cite}
\usepackage{textcomp}
\usepackage{amssymb}
\usepackage{latexsym}
\usepackage[inline]{enumitem}
\usepackage{rotating}
\usepackage{mathtools}
\usepackage{xcolor}
\usepackage{supertabular}
\usepackage{subcaption}

\usepackage[textsize=tiny, textwidth=1.4cm]{todonotes}
\def\BibTeX{{\rm B\kern-.05em{\sc i\kern-.025em b}\kern-.08em
    T\kern-.1667em\lower.7ex\hbox{E}\kern-.125emX}}

\begin{document}

\title{Online Local Boosting: improving performance in online decision trees\\
\thanks{This study was financed in part by the Coordenação de Aperfeiçoamento de Pessoal de Nível Superior - Brasil (CAPES) - Finance Code 001, Coordination for the National Council for Scientific and Technological Development (CNPq) of Brazil - Grant of Project 420562/2018-4, Fundação Araucária (Paraná, Brazil),  and Fundação de Amparo à Pesquisa do Estado de São Paulo (FAPESP) grant 2018/07319-6.}
%\thanks{This study was financed in part by the Coordenação de Aperfeiçoamento de Pessoal de Nível Superior - Brasil (CAPES) - Finance Code 001 and Fundação de Amparo à Pesquisa do Estado de São Paulo (FAPESP) grant 2018/07319-6.}
}

% \author{\IEEEauthorblockN{1\textsuperscript{st} Author 1}
% \IEEEauthorblockA{\textit{Department 1} \\
% \textit{University 1}\\
% xxxxx \\
% xxxxx}
% \and
% \IEEEauthorblockN{2\textsuperscript{nd} Author 2}
% \IEEEauthorblockA{\textit{Department x} \\
% \textit{University x}\\
% xxxxx \\
% xxxxx}
% \and
% \IEEEauthorblockN{3\textsuperscript{rd} Author 3}
% \IEEEauthorblockA{\textit{Department x} \\
% \textit{University x}\\
% xxxxx \\
% xxxxx}
% \and
% \IEEEauthorblockN{4\textsuperscript{th} Author 4}
% \IEEEauthorblockA{\textit{Department x} \\
% \textit{University x}\\
% xxxxx \\
% xxxxx}
% }

\author{\IEEEauthorblockN{1\textsuperscript{st} Victor G. Turrisi da Costa}
\IEEEauthorblockA{\textit{Computer Science Department} \\
\textit{Londrina State University}\\
Londrina, Brazil \\
victorturrisi@uel.br}
\and
\IEEEauthorblockN{2\textsuperscript{nd} Saulo Martiello Mastelini}
\IEEEauthorblockA{\textit{Institute of Mathematics and Computer Sciences} \\
\textit{University of S{\~a}o Paulo}\\
S{\~a}o Carlos, Brazil \\
mastelini@usp.br}
\and
\IEEEauthorblockN{3\textsuperscript{rd} André C. Ponce de Leon Ferreira de Carvalho}
\IEEEauthorblockA{\textit{Institute of Mathematics and Computer Sciences} \\
\textit{University of S{\~a}o Paulo}\\
S{\~a}o Carlos, Brazil \\
andre@icmc.usp.br}
\and
\IEEEauthorblockN{4\textsuperscript{th} Sylvio Barbon Jr.}
\IEEEauthorblockA{\textit{Computer Science Department} \\
\textit{Londrina State University}\\
Londrina, Brazil \\
barbon@uel.br}
}

% \and
% \and

\maketitle

\maketitle

\begin{abstract}
As more data are produced each day, and faster, data stream mining is growing in importance, making clear the need for algorithms able to fast process these data.
Data stream mining algorithms are meant to be solutions to extract knowledge online, specially tailored from continuous data problem.
Many of the current algorithms for data stream mining have high processing and memory costs. Often, the higher the predictive performance, the higher these costs.
To increase predictive performance without largely increasing memory and time costs, this paper introduces a novel algorithm, named Online Local Boosting (OLBoost), which can be combined into online decision tree algorithms to improve their predictive performance without modifying the structure of the induced decision trees.
For such, OLBoost applies a boosting to small separate regions of the instances space.
Experimental results presented in this paper show that by using OLBoost the online learning decision tree algorithms can significantly improve their predictive performance.
Additionally, it can make smaller trees perform as good or better than larger trees.

\end{abstract}


\section{Introduction}
The amount of data gathered from different sources is growing exponentially. As these data usually come in streams, there is a need for fast, reliable, and incremental data stream mining algorithms \cite{Domingos2000, Hoens2012, COSTA201822}.
These algorithms have to deal with new challenges, like concept drift (CD), time and memory constraints, and well-known problems like class imbalance and overfitting.
Many efforts have been made trying to tackle these challenges.
To increase predictive performance and deal with CD, ensemble learners have been employed with high success in data stream scenarios \cite{Oza2005, KRAWCZYK2017132}, however at the cost of requiring more computational resources.
On the other hand, although using a single classifier demands fewer resources, their predictive performance is usually lower and also are prone to failing to adapt and detect CD.
Hence, a solution that increases the predictive performance without severely impacting computational resources is highly desirable \cite{Domingos2000, Gama2003, Gama2010, KRAWCZYK2017132}.

% DRAFT
Online solutions can fail to detect important aspects of the data, e.g. instances lying in decision boundaries. This is due to the fact they only process each instance once, differently from batch solutions that exploit additional information from data distributions to improve prediction performance. The well-known Boosting ensemble approach is built upon this idea: particularly concentrate on the challenging examples. AdaBoost~\cite{dietterich2000ensemble}, a boosting algorithm, assigns different weights for the samples, giving more importance for the cases that are harder to predict. The same reasoning could be applied in streaming scenarios. Our idea is to focus on difficulty samples more than once when updating the learning models. Hence, reducing the needed amount of time for learning potentially complex patterns. Besides achieving better predictive performance, this approach does not bring significant impacts on the learning time and memory resources.

Taking into consideration the aspects discussed before, in this paper we present a novel algorithm for data stream classification, named \textit{Online Local Boosting} (OLBoost).
The OLBoost, here proposed for online decision trees (ODT), works inside each leaf to dynamically adjust the incoming instances weights towards increasing the predictive performance.
It works in parallel with the online decision tree inducing algorithm and does not interfere with the decision tree induction algorithm, is used solely to predict new incoming instances. Our proposal performs local boosting in the sense that only the leaf predictors are boosted towards increasing predictive performance.
This work assesses the impact of OLBoost using eleven benchmark datasets.
For such, it assess the impact of OLBoost inside the Very Fast Decision Tree (VFDT) \cite{Domingos2000} and the Strict VFDT (SVFDT) \cite{COSTA201822}. 
Experimental results showed that, when coupled with ODT, OLBoost improves accuracy without high overheads in time and memory costs.

The paper is organised as follows: Section~\ref{sec:related} presents works related to our proposal. Section~\ref{sec:odt} gives some background on ODT building, including a brief description of VFDT and SVFDT. Following, Section~\ref{sec:olboost} presents our proposal. Our experimental setup is detailed in Section~\ref{sec:setup}. We discuss the obtained results in Section~\ref{sec:results} and present our final considerations and some venues for future research in Section~\ref{sec:final}.

\section{Related Work}\label{sec:related}

Many techniques have been proposed to increase the predictive performance of ODTs.
They can be divided into three main groups: structural modification of the decision tree, additional prediction strategies with the same structure; and ensembles.

Modifying the structure of the decision tree is a very well-explored area.
The Concept-adapting VFDT (CVFDT) algorithm, proposed by Hulten et al. \cite{Hulten2001}, keeps secondary trees in memory and constantly assess these trees to check if they have a higher predictive performance than the original tree and uses a sliding window to discard outdated instances.
The Hoeffding Option Tree (HOT) proposed by Pfahringer et al. \cite{Pfahringer2007} introduces the concept of option nodes instead of normal split nodes.
An option node is essentially a split node which tests for multiple conditions at the same time.
When a new instance arrives at an option node, it travels along with all children nodes where the conditions are true.
Predictions are done by averaging all paths.
These techniques generally revolve around building a larger decision tree or using complex algorithms to discard outdated information, which also impacts computational costs.

The second group, which uses additional prediction strategies in the leaves is where our work is situated. Gama et al. \cite{Gama2003} first introduced the idea of functional leaves to increase the predictive performance of the VFDT.
These leaves use a Naive Bayes (NB) or an Adaptive NB (ANB) algorithm to further increase predictive performance \cite{Gama2003}.
To the best of our knowledge, no further works explore other solutions which belong to this group.

Lastly, many ensembles have been proposed. Oza \cite{Oza2005} first adapted bagging and boosting to the online scenario with the OzaBagging and OzaBoosting ensembles. Bifet et al. \cite{Bifet2010} improved OzaBagging with Leveraging Bagging (LevBag) by adding an ADWIN to monitor the error of each base learner and increasing the variability of the instances' weights when performing bagging. Adaptive Random Forests, proposed by Gomes et al. \cite{Gomes2017}, further improved LevBag with ideas from the Random Forest algorithm. Online Accuracy Updated Ensemble \cite{BRZEZINSKI201450} uses a sliding window to maintain a set of weighted base learners. All these algorithms offered some increase in predictive performance. Nonetheless, this is accompanied by a great increase in computational (memory and time) costs without a sufficient increase in performance to justify \cite{CostaMCB18}. 
The most recent studies addressing ODT predictive improvements were focused on the last group, ensembles solution. However, we believe that time and memory cost could be slightly affected when improving the predictive performance by exploring a boosting approach in the prediction strategy without modifying the ODT induction.

\section{Online Decision Trees}\label{sec:odt}

Two ODT algorithms were used in the experiments carried out in this work, VFDT and its recent variation, which focus on reducing memory costs, SVFDT.

\subsection{Very Fast Decision Tree}

The VFDT \cite{Domingos2000} is a tree-based ML algorithm for data streams based on the Hoeffding Bound (HB) theorem.
When growing a tree, the VFDT employs the HB to perform a node split.
After evaluating the candidate features at a split attempt with a heuristic measure $G(.)$ (e.g., Information Gain (IG) or Gini Index (GI)), VFDT uses the HB theorem to check whether the best split candidate would remain the best if the tree received additional instances.

VFDT keeps and updates the instances class distribution in a vector at each leaf to count the number of instances from each class.
Likewise, counting procedures and numerical estimators are also employed to maintain the relationship between the feature values and class distributions.
By doing so, VFDT can induce a model from a single instance at a time using limited computational memory resources.
Additionally, under realistic assumptions, it has the same asymptotic performance as the induction of a decision tree by a standard batch algorithm \cite{Gama2010}.

Finally, VFDT has a hyperparameter $\tau$ to support tree growth when features have similar $G(.)$ values; uses an Adaptive Naive Bayes (ANB) \cite{Gama2003} at leaves to increase predictive performance; and uses a $GP$ hyperparameter that defines the amount of instances needed by each leaf between split attempts.

\subsection{Strict Very Fast Decision Tree}

The SVFDT algorithm \cite{COSTA201822} modifies the VFDT to create smaller decision trees while maintaining predictive performance.
SVFDT has two different versions, SVFDT-I and SVFDT-II, both following the assumptions that:
\begin{enumerate}
    \item A leaf node should only split if there is a minimum uncertainty of class assumption associated with the instances, according to previous and current statistics (i.e., a high entropy).
    \item All leaf nodes should observe a similar number of instances to be turned into split nodes.
    \item The feature used for splitting should have a minimum relevance according to previous statistics (i.e., a high IG).
\end{enumerate}

The SVFDT applies additional rules to hold tree growth using the following $\varphi$ function:
\begin{equation*}
    \varphi(x, X) =
    \begin{dcases}
        \text{True},    & \text{if } x \geq \overline{X} - \sigma(X) \\ 
        \text{False},   & \text{otherwise}
    \end{dcases}
\end{equation*}
where $X$ is a set of observed values, $\overline{X}$ is their mean, $\sigma(X)$ is their standard deviation, and $x$ is a new observation.

First, consider that statistics computed at the time a leaf satisfy the splits conditions of the  VFDT (according to the $HB$ or $\tau$) are marked with a \textit{satisfiedVFDT}.
For example, the $i$-th that that this occurred, the entropy of that leaf would be marked as $H_{\text{satisfiedVFDT}_i}$, the $G(.)$ value of the best feature would be $G_{\text{satisfiedVFDT}_i}$, and the number of instances seems $n_{\text{satisfiedVFDT}_i}$.

The SVFDT splits a leaf when it satisfies all the conditions imposed by the VFDT and four additional constraints applied to each leaf $l$:
%Them being: 
\begin{enumerate}
    \item $\varphi(H_l, \{H_{l_0}, ..., H_{l_L}\})$,
    \item $\varphi(H_l, \{H_{\text{satisfiedVFDT}_0}, ..., H_{\text{satisfiedVFDT}_S}\})$,
    \item $\varphi(G_{l}, \{G_{\text{satisfiedVFDT}_0}, ..., G_{\text{satisfiedVFDT}_S}\})$,
    \item $n\textsubscript{\textit{l}} \geq \overline{\{n_{\text{satisfiedVFDT}_0}, ..., n_{\text{satisfiedVFDT}_S}\}}$,
  \end{enumerate}
where $L$ is the total number of leaves and $S$ is the total number of split attempts that satisfied the VFDT constraints.

The SVFDT-II uses additional skipping mechanisms to speed-up growth when class uncertainty is too high or this uncertainty is largely reduced.
It employs the following $\varpi$ function:
\begin{equation*}
    \varpi(x, X) =
    \begin{dcases}
        \text{True},    & \text{if } x \geq \overline{X} + \sigma(X) \\ 
        \text{False},   & \text{otherwise}
    \end{dcases}
\end{equation*}
To check if either
\begin{enumerate}
    \item $\varpi(H_l, \{H_{\text{satisfiedVFDT}_0}, ..., H_{\text{satisfiedVFDT}_S}\})$ or
    \item $\varpi(G_{l}, \{G_{\text{satisfiedVFDT}_0}, ..., G_{\text{satisfiedVFDT}_S}\})$
  \end{enumerate}
hold true, and then ignoring all the other $\varphi$ constraints.

\section{Online Local Boosting}\label{sec:olboost}

The Online Local Boosting (OLBoost) is a simple algorithm designed to increase predictive performance, primarily when working coupled with ODTs.
It is based on the assumption that to increase predictive performance, instances being wrongly classified are required to be used more times when inducing a model.
On the other hand, instances that are easily classified can even be dis-considered during the training phase.
However, the application of this procedure may introduce some pitfalls (e.g., overfitting), since it changes the original data stream distribution.
To deal with this pitfalls, OLBoost works independently from tree growth, empirically reducing this problem. More specifically, OLBoost works as the predictor of each leaf in a tree. However, it must be observed that the OLBoost can use as its core either a most common (MC) prediction, NB or ANB. Consequently, according to the strategy used different statistics about the instances need to be stored.
For MC, a simple class distribution is sufficient, while for NB or ANB, it needs additional statistics about the nominal and numerical feature distributions.
These additional statistics are computed in the same way as a simple VFDT using as leaf predictor the ANB.

\begin{figure*}[!ht]
\begin{center}
\includegraphics[width=0.8\textwidth, keepaspectratio]{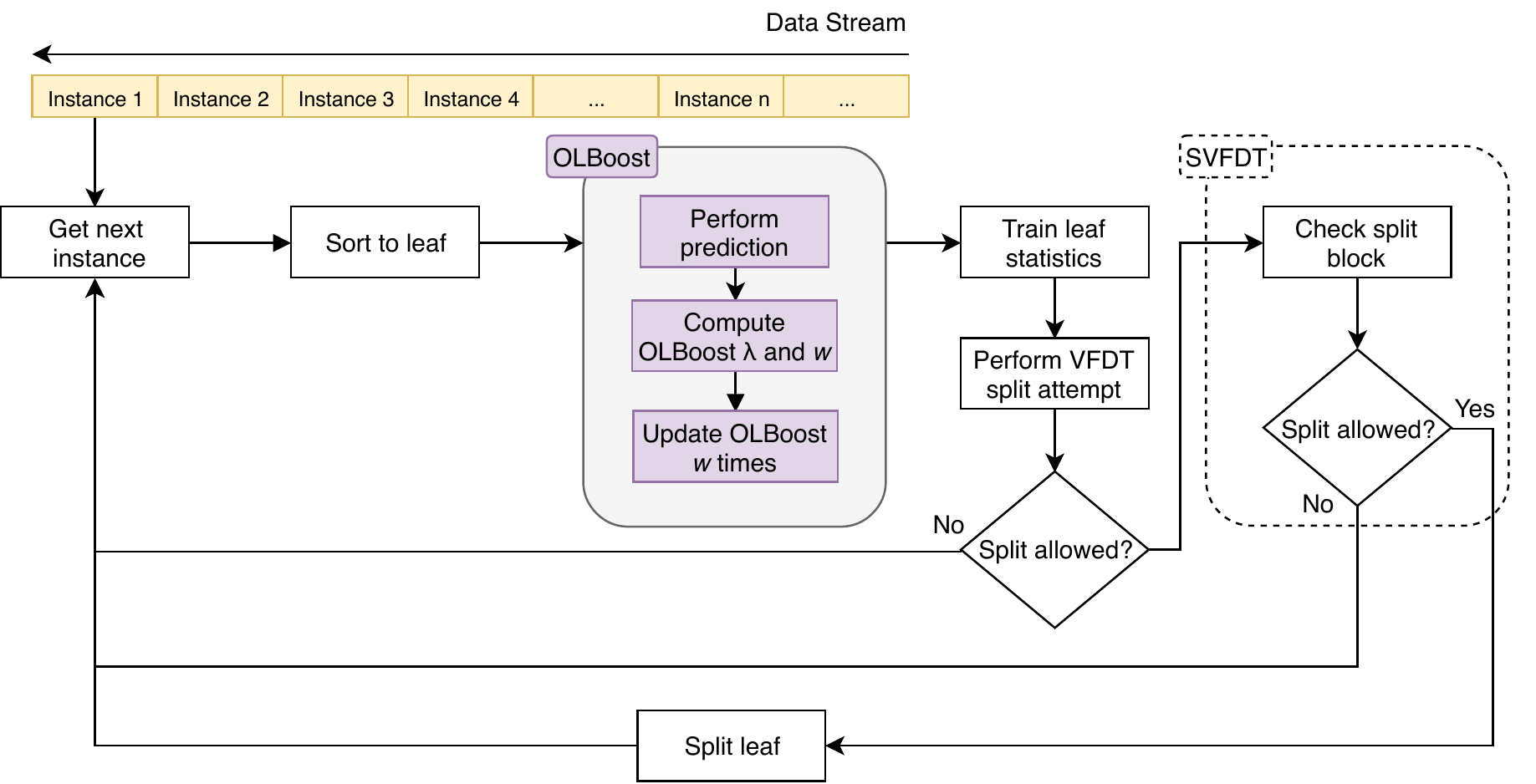}
\end{center}
% \vspace{-16pt}
\caption{OLBoost main steps using 
%overview with 
VFDT/SVFDT in the training and prediction phases.}
\label{fig:diagram-olboost}
\end{figure*}

Fig. \ref{fig:diagram-olboost} describes how OLBoost works. Since we evaluate in this work two versions of OLBoost when coupled with VFDT and SVFDT, these couplings are illustrated in the figure.
Both training and prediction phases are presented in the figure. Nonetheless, the presented scheme could be easily adapted to other ODT algorithms.

The instances of an (unbounded) data stream are presented one at a time to the online learning algorithm.
Each instance is defined as $i=(\mathbf{x}, y)$, where $\mathbf{x}$ is the feature vector and $y$ the real class.
First, an instance $i$ is presented to the algorithm and sorted to a leaf in the ODT.
This process corresponds to traversing the tree according to its split nodes until $i$ arrives at a leaf.
Then, OLBoost tries to predict this instance class, outputting a probabilities vector which is stored. 
When using NB or ANB as the core of OLBoost, the probabilities outputted by these algorithms are not true probabilistic representations (between 0 and 1 and with a sum of 1).
To deal with this, we tested both softmax or simply diving each probability by the sum of all probabilities and found out that the latter yielded better results.

Afterwards, the boosting procedure is performed in a similar manner as described in \cite{Oza2005}, by sampling weights from a Poisson distribution.
It works by computing a $\lambda$ variable that is then used to select the instance weight $w$.
Note that $w$ is equivalent to the number of times that an instance is used to update the OLBoost. Our proposal computes the instance weights by linearly combining a range of allowed Poisson distribution parameters and the probability outputted by the ODT model.
In this sense, to compute $\lambda$ the following equation is used:
\begin{equation*}
\lambda = (min_\lambda - max_\lambda) * P(y) + max_\lambda,
\end{equation*}
where $min_\lambda$ and $max_\lambda$ are hyperparameters which delimit the minimum and maximum possible values for $\lambda$ and $P(y)$ corresponds to the probability estimated by the leaf of $i$ being from its real class.
Then, $w$ is drawn by sampling from the Poisson($\lambda$) distribution.
Note that $\lambda$ and $P(y)$ are inversely proportional, i.e., as the prediction becomes more accurate, $\lambda$ decreases, which in turn statistically reduces the chance of $w$ being a larger value.
Lastly, the statistics used by the predictor inside the OLBoost (MC, NB or ANB) are updated $w$ times with $i$.
After this, the OLBoost does not interfere with tree growth.
The instance is then used to update the leaf statistics with a normal weight of 1.
Then, a split attempt if performed.

Considering the time costs of using OLBoost, we have some different scenarios.
If we compare using it with using a MC or NB predictor, then the OLBoost has time costs associated with making a prediction (which will vary according to the predictor inside the OLBoost), computing $\lambda$, sampling $w$ and updating its statistics.
On the other hand, if we compared against using an ANB, then there would be no additional prediction costs since ANB has to perform a prediction to see which predictor (MC or NB) is performing better.
Memory costs will increase according to the predictor used inside the OLBoost, being more costly when using a NB or ANB.
However, when using ensembles to improve prediction performance, both memory cost increase around $m$ times
When the models are sequentially updated, the processing cost also increases by around $m$ times, where $m$ is the number of models inside the ensemble.
Nonetheless, when predictive performance needs to be maximised and the other resources are available, OLBoost coupled with an ODT can be used in ensembles.

\section{Experimental setup}\label{sec:setup}

To evaluate the impact of the OLBoost in the VFDT and SVFDTs, eleven benchmark datasets, commonly used in data stream mining experiments,  were selected: \textbf{agrawal}, \textbf{cover\_type}, \textbf{electricity}, \textbf{hyper}, \textbf{led24}, \textbf{poker}, \textbf{rbf} (with 500k and 10 features, 1M instances and 10 features, and 250k instances and 50 features), \textbf{sea} and \textbf{usenet}. In all the cases, the prequential evaluation scheme was employed for evaluating the algorithms~\cite{Gama2010}. %\todo{Saulo diz: Que tal uma tabela com as caracteristicas basicas dos datasets?}

The VFDT and SVFDTs used ANB and $GP$ and $\tau$ were varied: $GP \in (100, 200, 400, 800, 1000)$ and $\tau \in (0.01, 0.05, 0.10, 0.15, 0.20)$.
Likewise, $min_\lambda$ and $max_\lambda$ were varied, with performance increasing as $max_\lambda$ increased.
Based on empirical evaluations the best all around results plateaued for around $min_\lambda = 1$ and $max_\lambda = 12$, hereby recommended as default values.

We evaluated the compared algorithms in terms of predictive performance, running time, and the amount of required memory. The predictive performance was measured in terms of accuracy. We accounted for the total running time of the algorithms in seconds and measured the final size of the models (in MB) at the end of the data streams.

All algorithms were implemented in Python 3.7\footnote{https://www.python.org/} and Cython\footnote{http://cython.org/}, a superset of the Python language that allows code to be written in Python and compiled to C extensions, and are available online at \footnote{https://github.com/vturrisi/pystream}.

\section{Results and Discussions}\label{sec:results}

\begin{figure*}[ht!]
    \begin{center}
    \includegraphics[angle=0,origin=c, width=0.9\textwidth, keepaspectratio]{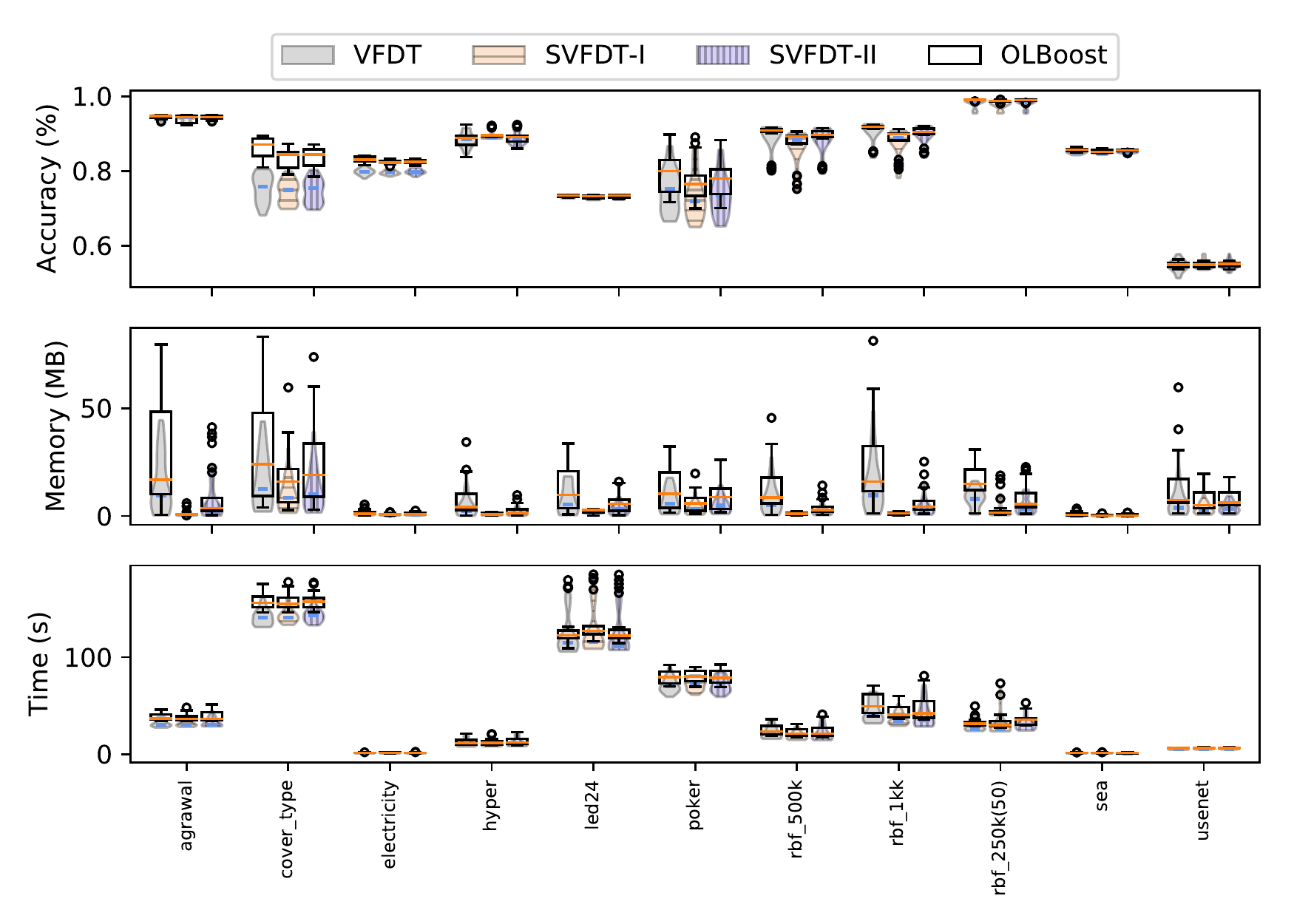}
    \end{center}
    \vspace{-20pt}
\caption{Boxplot and violin plot of the performance metrics for the VFDT, SVFDT-I and SVFDT-II with and without OLBoost.}
\label{fig:boxplot}
\end{figure*}

Fig. \ref{fig:boxplot} presents the boxplots and violin plots of the performance metrics for each dataset separately while varying $GP$ and $\tau$.
The violins represent the algorithms without using OLBoost, while the overlapped boxplots represent the same algorithm with OLBoost. As previously discussed, the values of $min_\lambda$ and $max_\lambda$ were kept constant during the experiments.

Considering accuracy, it is possible to see that excluding \textbf{agrawal} and \textbf{usenet} datasets, using OLBoost increases the median performance. Moreover, OLBoost outperforms the traditional ODT variants in all hyperparameter configurations for the \textbf{cover\_type} and \textbf{electricity} datasets.
While the traditional ODT algorithm can reach similar predictive performance in some datasets by selecting high values for $\tau$ (creating a deeper tree), using OLBoost grants higher or very similar performance while using a much shallower tree.

Concerning memory usage, OLBoost increases consumption in all cases, but this was expected since it essentially doubles the memory sizes of the leaves. This fact can, however, be neglected for most of the applications, given that our proposal reaches higher predictive performances than standard ODT algorithms. OLBoost can also use shallower structures while maintaining competitive accuracy. In the future, we intend to evaluate new strategies for tree growth stopping based on the measured accuracy. This fact could enhance the advantages of our proposal against other algorithms. 

For time costs, a small variation is perceived, but without an impacting increase. Given the previously discussed aspects, settings which lead OLBoost to structures with smaller depths would be preferred to decrease time processing costs. The robustness of our proposal accuracy-wise enables this kind of balance between size and performance. Hence, we advise using OLBoost while tuning ODT's hyperparameters to create smaller trees maximising the efficiency between predictive performance and memory consumption.

It is also worth mentioning the SVFDT variants when coupled with OLBoost were able to obtain predictive performance comparable to those of VFDT. Nonetheless, we limited our analysis for pairwise comparison of tree settings, i.e., we compared the performance of the same hyperparameter set using or not OLBoost. An interesting venue for the future evaluation would be comparing the performance obtained by SVFDT with OLBoost against the traditional VFDT algorithm. This could lead to more accurate decision models that are restrictive in the usage of computational resources.

\begin{figure}[ht!]
\begin{center}
\includegraphics[width=0.48\textwidth, keepaspectratio]{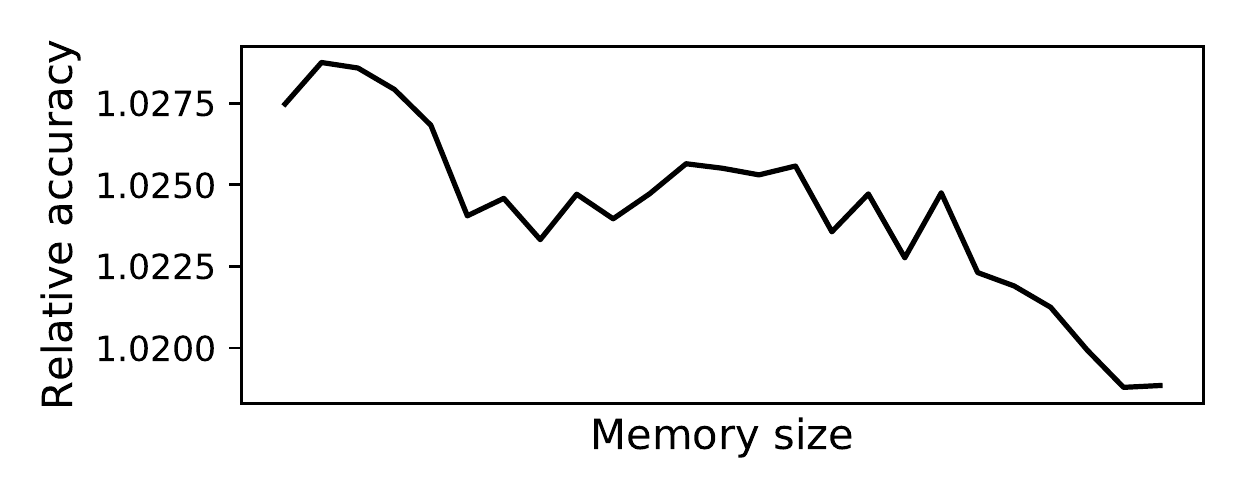}
\end{center}
\vspace{-20pt}
\caption{Average relative accuracy considering all trees when using OLBoost sorted by tree size.}
\label{fig:rel_acc}
\end{figure}

We also evaluated the benefits of using OLBoost as the tree structures grow.
Fig. \ref{fig:rel_acc} presents the average relative accuracy considering all trees when using OLBoost sorted by tree size. The relative accuracy is calculated by dividing the accuracy of the OLBoost-based models by the accuracy of their standard counterparts. Therefore, relative accuracies greater than $1$ implies that our proposal led to higher accuracies, whereas the contrary holds true to values below $1$.
As the trees increase, the gains obtained by OLBoost become smaller. We expected such behaviour since the prediction models become more specialised in the incoming concepts as they process more examples. However, considering all ODTs evaluated and the datasets, OLBoost always presents a gain in accuracy.

Lastly, the statistical difference between the six algorithm combinations (VFDT, SVFDT-I, SVFDT-II, and their OLBoost variants: O\_VFDT, O\_SVFDT-I, and O\_SVFDT-I) was assessed using the Wilcoxon signed-rank test \cite{wilcoxon1992individual}.
We used the results obtained by each pair of algorithms, considering all the evaluated hyperparameters.
Thus, we used a sample size of $25$ (the number of different evaluated hyperparameter combinations per dataset) for each performed test.
First, a two-sided test was computed to verify whether the median of the differences between each algorithm pair was zero (null hypothesis) or not (alternative hypothesis).
In case the alternative hypothesis was observed, an additional one-sided test was performed to verify whether the median of the differences was positive or negative. Therefore, we can state for each algorithm pair $a$ and $b$, the number of datasets $a$ was statistically better than $b$, and vice versa. We present the observed results in Tables \ref{tab:wilcoxon_mean_acc}, \ref{tab:wilcoxon_memory}, and \ref{tab:wilcoxon_time}.

Considering accuracy (Table \ref{tab:wilcoxon_mean_acc}), the O\_VFDT was statistically better than the other algorithms more times than any other algorithm.
Moreover, we can also observe that by adding OLBoost to the SVFDT-I and SVFDT-II, both algorithms were able to surpass the accuracy achieved by even the best performing technique (VFDT).
From a memory perspective (Table \ref{tab:wilcoxon_memory}) it is possible to see that adding OLBoost to any of the algorithms statistically increases memory costs.
Nonetheless, O\_SVFDT-I and O\_SVFDT-II were still able to statistically consume less memory than the VFDT in six and five datasets, respectively.
Considering time (Table \ref{tab:wilcoxon_time}), OLBoost always increases computational costs regardless of the base ODT.
Despite being slower, the real difference values are around 10 seconds throughout processing the whole stream. This amount can be neglected depending on the type of data stream application.

%\cite{demvsar2006statistical}.
%The tests were performed considering the results for each dataset-hyperparameter pair as an observation for each algorithm.
%Critical Distances (CD) diagrams for the Nemenyi test are presented in Figs \ref{fig:nemenyi_acc}, \ref{fig:nemenyi_mem} and \ref{fig:nemenyi_time}.
%Groups that are not significantly different ($\alpha$= 0.05 and CD = 0.45) are connected.

%Considering accuracy (Fig. \ref{fig:nemenyi_acc}), the only algorithms statistically equivalent were OLBoost coupled with the SVFDT-I and the VFDT. The algorithms coupled with OLBoost outperformed the versions without it in all the cases. For memory (Fig. \ref{fig:nemenyi_mem}), OLBoost increases memory consumption as previously discussed.
%Nonetheless, when used inside a SVFDT-I tree, it is still statistically equivalent to the SVFDT-II without OLBoost. This again reinforces the possibility of obtaining accurate models without compromising memory usage.
%Lastly, from the time perspective (Fig. \ref{fig:nemenyi_time}), using OLBoost increases time costs regardless of the ODT variant.
%Despite being slower, the real difference values are around 10 seconds throughout processing the whole stream. This amount can be neglected depending on the type of data stream application.

% \begin{figure}[ht!]
% \center
% \subfigure[ref1][Accuracy]{
% \input{ESANN_OLBoost_ACC.tex}
% }
% \qquad
% \subfigure[ref2][Memory]{
% \input{ESANN_OLBoost_MEM.tex}
% }
% \qquad
% \subfigure[ref2][Time]{
% \input{ESANN_OLBoost_TIME.tex}
% }

% \caption{\small{Critical distance diagram for Nemenyi test.}}

% \end{figure}

\begin{table}[!htbp]
  \centering
  \caption{Wilcoxon signed-rank test results with $\alpha = 0.05$: each cell shows the number of times the algorithm in the row was statistically better than the algorithm in the column, regardless the hyperparameters used.}
  \begin{subtable}{0.48\textwidth}
    \centering
    \caption{Mean Accuracy}
    \label{tab:wilcoxon_mean_acc}
    \resizebox{\textwidth}{!}{
    \begin{tabular}{|c|c|c|c|c|c|c|}
      \hline
      \textbf{Algorithm/Algorithm}     & \textbf{O\_SVFDT-I} & \textbf{O\_SVFDT-II} & \textbf{O\_VFDT} & \textbf{SVFDT-I} & \textbf{SVFDT-II} & \textbf{VFDT} \\ \hline
      \textbf{O\_SVFDT-I}  & --                & 1                 & 1             & 9       & 7        & 7    \\ \hline
      \textbf{O\_SVFDT-II} & 6                & --                 & 1             & 8       & 9        & 7    \\ \hline
      \textbf{O\_VFDT}     & 9                & 8                 & --             & 9       & 8        & 8    \\ \hline
      \textbf{SVFDT-I}           & 0                & 0                 & 0             & --       & 1        & 1    \\ \hline
      \textbf{SVFDT-II}          & 0                & 0                 & 0             & 5       & 0        & 0    \\ \hline
      \textbf{VFDT}              & 3                & 2                 & 1             & 7       & 6        & --    \\ \hline
    \end{tabular}}
  \end{subtable}
  \begin{subtable}{0.48\textwidth}
    \centering
    \caption{Memory size}
    \label{tab:wilcoxon_memory}
    \resizebox{\textwidth}{!}{
    \begin{tabular}{|c|c|c|c|c|c|c|}
      \hline
      \textbf{Algorithm/Algorithm}     & \textbf{O\_SVFDT-I} & \textbf{O\_SVFDT-II} & \textbf{O\_VFDT} & \textbf{SVFDT-I} & \textbf{SVFDT-II} & \textbf{VFDT} \\ \hline
      \textbf{O\_SVFDT-I}  & --                & 10                & 11            & 0       & 0        & 6    \\ \hline
      \textbf{O\_SVFDT-II} & 0                & --                 & 11            & 0       & 0        & 5    \\ \hline
      \textbf{O\_VFDT}     & 0                & 0                 & --             & 0       & 0        & 0    \\ \hline
      \textbf{SVFDT-I}           & 10               & 11                & 11            & --       & 10       & 11   \\ \hline
      \textbf{SVFDT-II}          & 6                & 11                & 11            & 0       & --        & 11   \\ \hline
      \textbf{VFDT}              & 2                & 4                 & 11            & 0       & 0        & --    \\ \hline
    \end{tabular}}
  \end{subtable}
  \begin{subtable}{0.48\textwidth}
    \centering
    \caption{Time (s)}
    \label{tab:wilcoxon_time}
    \resizebox{\textwidth}{!}{
    \begin{tabular}{|c|c|c|c|c|c|c|}
      \hline
      \textbf{Algorithm/Algorithm}     & \textbf{O\_SVFDT-I} & \textbf{O\_SVFDT-II} & \textbf{O\_VFDT} & \textbf{SVFDT-I} & \textbf{SVFDT-II} & \textbf{VFDT} \\ \hline
      \textbf{O\_SVFDT-I}  & --                & 3                 & 3             & 0       & 0        & 0    \\ \hline
      \textbf{O\_SVFDT-II} & 1                & --                 & 2             & 0       & 0        & 0    \\ \hline
      \textbf{O\_VFDT}     & 1                & 2                 & --             & 0       & 0        & 0    \\ \hline
      \textbf{SVFDT-I}           & 11               & 11                & 11            & --       & 1        & 2    \\ \hline
      \textbf{SVFDT-II}          & 11               & 11                & 11            & 1       & --        & 3    \\ \hline
      \textbf{VFDT}              & 11               & 11                & 11            & 2       & 4        & --    \\ \hline
    \end{tabular}}
  \end{subtable}
\end{table}

\section{Conclusion and Future Work}\label{sec:final}

This work presented a novel algorithm to increase the predictive performance of ODTs for data stream mining. We considered a large and varied set of benchmark datasets to compare our proposal against traditional ODT algorithms.
According to the experimental results, OLBoost is able to increase accuracy performance.
Additionally, OLBoost produced trees shallower and with better predictive performance better than the other algorithms, reducing the memory needed, speeding up the test and potentially improving the model's interpretability. Reducing model size also helps to combat overfitting. In addition, we observed that variants of SVFDT with OLBoost were capable of presenting predictive performance values competitive to those obtained by the traditional VFDT.
As future work, we intend to evaluate the possibility of using OLBoost in a restricted set of leaves to increase predictive performance while reducing memory costs. Moreover, the evaluation of OLBoost in ensemble algorithms can also be explored. The adaptation of OLBoost to regression tasks is another possible venue for future research. Finally, our proposal can be extended to other online prediction algorithms, such as k-Nearest Neighbours.

% Já tem o thanks na primeira página

\scriptsize
\bibliographystyle{unsrt}
\bibliography{bibtex}

%\end{footnotesize}

% ****************************************************************************
% END OF BIBLIOGRAPHY AREA
% ****************************************************************************

\end{document}